# Confiding in and Listening to Virtual Agents: The Effect of Personality

**Jingyi Li**[1]
Univ. of California, Berkeley
jli@juji-inc.com

**Michelle X. Zhou**
Juji, Inc.
{mzhou, hyang}@juji-inc.com

**Huahai Yang**
Juji, Inc.

**Gloria Mark**
Univ. of California, Irvine
gmark@uci.edu

## ABSTRACT
We present an intelligent virtual interviewer that engages with a user in a text-based conversation and automatically infers the user's psychological traits, such as personality. We investigate how the personality of a virtual interviewer influences a user's behavior from two perspectives: the user's willingness to confide in, and listen to, a virtual interviewer. We have developed two virtual interviewers with distinct personalities and deployed them in a real-world recruiting event. We present findings from completed interviews with 316 actual job applicants. Notably, users are more willing to confide in and listen to a virtual interviewer with a serious, assertive personality. Moreover, users' personality traits, inferred from their chat text, influence their perception of a virtual interviewer, and their willingness to confide in and listen to a virtual interviewer. Finally, we discuss the implications of our work on building hyper-personalized, intelligent agents based on user traits.

## INTRODUCTION
From human capital management to healthcare, interviewing is often used to obtain information from target respondents and to assess certain characteristics of the respondents. For example, a hiring manager may interview a pool of job applicants over the phone to discover their career interests and assess their characteristics, such as personality and motivations, before determining their fit for the position.

While traditional interviewing by people helps gather more and deeper information, it presents several limitations. First, it is difficult to scale since a human interviewer can afford to conduct only a limited number of interviews per day. Second, much research shows that a human interviewer's personal factors, such as personality, mood, or biases, can affect interview results (e.g., [19, 50]). Third, respondents may not feel comfortable disclosing sensitive information to human interviewers due to social desirability biases [27].

To overcome the limitations of a human interviewer while providing a personal touch, we have built a *virtual interviewer*, an intelligent agent that interviews a user through a text-based conversation. We call such a virtual interviewer a REP (Responsible, Empathetic Persona). As shown in Figure 1, not only does a REP (Kaya) conduct a text-based virtual interview with a user (James), but also *automatically* infers the user's traits, such as personality, based on the user's chat text given during the interview. Our decision to support a text-only conversation is based on two reasons. First, compared to the use of richer modalities like speech or gestures, a text-only interaction can be supported more robustly, suiting our goal of deploying a REP for real world use. Moreover, text-only interaction keeps a user focused on the interview without distractions [22, 39, 43].

Compared to a human interviewer, a REP has two distinct advantages. First, a REP does not experience fatigue and can scale to interview hundreds of thousands of users simultaneously. Second, a REP can render a more objective, unbiased assessment, since it does not bring personal emotions or human prejudice into an interview.

Since studies show that interviewers' personalities can directly affect interview outcome [19, 50], one of our challenges is to equip a REP with the "personality" of an effective virtual interviewer. This challenge is non-trivial for three reasons. First, existing research suggests that computers should take on *one* strong, consistent personality [37]. However, it is unclear which personality a REP should assume to achieve multiple interview goals. For example, a warm and cheerful personality may aid a REP in building rapport with its respondents, who can then relax and act authentically [2, 27]. On the other hand, a more serious and assertive REP may elicit more respect and better cooperation from the respondents [19, 36]. Second, most existing systems stress the use of both verbal and non-verbal cues to express an agent's personality [25, 40, 36]. However, it is unclear how a REP can effectively exhibit its personality through text-only expressions. Third, unlike most systems, which are used in a laboratory setting, our REP is used in the real world, including high-stake situations, such as job interviews. It is unknown how a REP's personality would affect users in such a context.

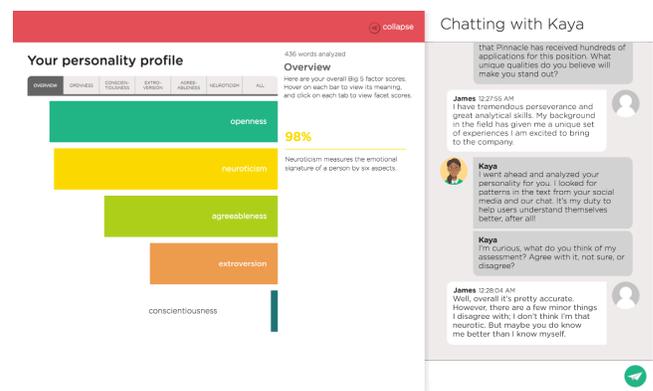

**Figure 1. Our text-based chat interface during an interview.**

To address the challenge mentioned above, as the first step, we decided to give each REP a strong, consistent personality as suggested by existing work [37]. We thus built two REPs with two distinct personalities, each aimed at achieving a key interview goal. One is a female persona (Kaya) with a warm and cheerful personality, while the other is a male persona (Albert) with a reserved and assertive personality. Based on existing work, we hypothesize that:

- **H1**: Users are more willing to confide in Kaya since she is better at building rapport with them;
- **H2**: Users are more willing to listen to Albert and follow his advice since he is better at gaining respect from them;
- **H3**: Individual differences of users, including their personality, impact their behavior with Kaya or Albert.

We deployed both Kaya and Albert to help an organization interview and pre-screen their job applicants. In this field deployment, 157 applicants finished their interview with Kaya, and 159 completed theirs with Albert. Our results reveal several new insights. First, the personality of a REP affects users' willingness to confide in and listen to the REP. Second, users' individual differences, in particular, their personality traits inferred by a REP, influence their perception of and behavior with the REP. As a result, our work presents two unique contributions: (1) novel, practical approaches to building effective virtual interviewers who can automatically assess users' characteristics; and (2) design implications for building hyper-personalized REPs based on our findings and automatically inferred user traits.

**RELATED WORK**

Our current work is influenced by studies on personification of agents and how the "personality" of machines influences user behavior. For example, studies show that equipping agents with human forms, such as faces, make the agents more likable and engaging [22, 39]. However, studies also find that it requires effort from users to interpret the meanings of human-like expressions [22], which may even create a distraction [48]. Moreover, users are less relaxed or assured when interacting with more human-like agents [39]. As one of our main goals is to create an atmosphere where users can relax and act authentically during a virtual interview, we decide to support a text-only conversation to minimize potential user distraction from interpreting human-like expressions of a virtual interviewer.

To understand how the personality of a computer may influence user behavior, there is much work on matching agent personality with users' personality and tasks. For example, Reeves and Nass [37] find that users favor a computer with a personality that matches their own. However, in human-robot interaction, studies show conflicting results. For example, Lee et al. [25] indicate that users enjoy interacting with a robot with a complementary personality, while Tapus et al. [40] show that users in rehabilitation therapy prefer to interact with a robot with a matching personality. Moreover, studies find that users expect the personality of an agent to match the task context, e.g., a serious agent for a serious task [36], which directly influences users' willingness to comply with the robot's instructions. While these studies provide insights into the effects of an agent's personality on users in various laboratory situations, it is unknown what personality an effective virtual interviewer should take on and how its personality would impact its users, especially in real-world, risk-bearing situations.

Another line of related work is a wide variety of efforts on developing embodied "virtual humans" (e.g., [2, 11, 20, 27, 27]). Closest to our work are efforts on creating 3D virtual interviewers or counselors who build rapport with users to encourage the disclosure of sensitive information [2, 27].

While we leverage findings in these works, such as the use of introductory questions [27] and reciprocal self-disclosure [2] to build rapport, and the use of different virtual personas to influence users' behavior [11, 20], our work differs in several aspects. First, although all these systems interpret user interaction behavior and maintain a user model, few focus on inferring a user's psychological traits as ours does. Second, most existing efforts focus on instrumenting the social skills of an embodied agent to build connections with users. In contrast, our work aims at equipping a REP with a strong, consistent personality via text communication to achieve multiple interview goals, such as eliciting authentic user input and cooperation.

Our work on automatically inferring a user's psychological characteristics, such as personality, from the user's communication text, is inspired by multiple efforts on studying the relationships between users' traits and their communication patterns. For example, research indicates that words or word categories in users' blogs and essays are correlated with self-reported or third-party-rated personality traits [30, 41, 49]. More recent studies show that users' vocabulary and behavior on social networks, such as Facebook, are also related to self-reported personality traits [21, 23]. Based on such findings, researchers have built various computational models to automatically infer one's personality traits from text (e.g., [1, 14, 29]). Like these computational models, ours also automatically infers a user's characteristics from text. However, unlike most of these models, which infer user traits based on the LIWC categories of word uses (e.g., [14, 29]) or high-level features (e.g., number of friends on Facebook [1]), our finer-grained computational model uses

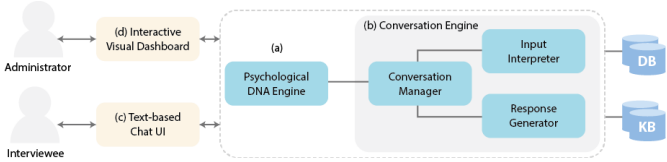

**Figure 2. System overview.**

rich linguistic cues (e.g., words, phrases, emoticons, and punctuations) to infer user traits.

## SYSTEM OVERVIEW AND KEY COMPONENTS

Figure 2 shows an overview of our virtual interviewer system. A REP conducts a virtual interview with a user (interviewee) through a text-based chat user interface (c), powered by two core engines: (a) psychological DNA engine and (b) conversation engine. The *psychological DNA engine* analyzes a user's interaction behavior and automatically infers the user's psychological characteristics, such as personality, strengths, and weaknesses. The *conversation engine* enables a REP to engage a user in a mixed-initiative conversation during a virtual interview. Similar to a human interviewer, a REP takes initiatives to guide the interview flow and solicit information from a user, while it allows the user to pose questions whenever appropriate. For example, a user may ask a clarification question or inquire about the position that s/he is applying for. Specifically, the *conversation manager* controls the conversation flow, the *input interpreter* processes user input, and the *response generator* outputs a REP's verbal utterances.

To help business administrators (e.g., hiring managers) harness the interview results and make decisions, we provide an interactive visual dashboard that summarizes interview results (d), shown in Figure 3. Our system also maintains several databases (DB) and knowledge bases (KB), which store various types of information, such as users' answers to various interview questions and inferred user traits.

Our system uses a web-based, client-server architecture. The server-side components are written in Clojure, while the client ones are written in HTML and JavaScript.

### Evidence-based Psychological DNA Engine

To automatically infer a user's traits from text, most existing approaches use the uncovered correlations between word categories (i.e., LIWC categories) and self-reported personality test scores (e.g., [1, 14, 29]). While these approaches have demonstrated their effectiveness in various applications (e.g., [26, 28, 29]), they have three main limitations. First, while using word categories is simple, the inference is coarse. This is because different words belonging to the same word category have the same weight, but their relationships to a trait may be quite different [49]. Moreover, there is a limited number of word categories (68 in LIWC), and many word categories may not show significant correlations with a trait. Second, the correlations are derived based on one's word use and self-reported personality test scores, which may not be completely truthful [51].

Third, these approaches consider only single words but miss out other important linguistic cues, such as phrases, and punctuations, which also reflect one's traits [21, 23].

To overcome these limitations, we have developed an evidence-based approach to inferring a user's traits from the user's text. While describing the approach in detail is out of the scope for this paper, we highlight the three main steps.

*Step 1: Gathering Trait-Related Evidence*
To build a trait inference model, we first gather training data—behavioral data generated by a person to capture this person's real-world activities that reflect associated characteristics. For example, an adventurous or creative person may have generated tweets or blog posts to reflect their adventurous or creative nature. Given a person's behavioral data, we then identify both manually and automatically trait-pertinent evidence, such as words, phrases, and punctuation marks. For example, for adventurous people, certain words or phrases would indicate their daring activities or excitement-seeking interests.

*Step 2: Mining Evidence-Trait Relationships*
We then mine the gathered data to extract various linguistic cues (evidence) and quantify their relationships with various traits. Linguistic cues may be words, phrases, punctuations, or emoticons, and serve as potential evidence to measure various traits.

To quantify the relationships between potential evidence and user traits, such as the Big 5 personality traits, we adopt a statistical inference approach, called Item Response Theory, a widely used approach in modern psychometrics [8]. In particular, we model the desired traits as latent factors and potential evidence as observed items. To estimate the relationships between any evidence $j$ and a trait $\theta$ for an individual $i$, we use the following formula:

$$X_{ij} = \mu_j + \lambda_j \theta_i + \varepsilon_{ij}; \; j=1, N \quad (1)$$

Here $X$ is the observed data containing $N$ pieces of evidence; $\mu_j$ is the occurrence rate of evidence $j$; $\lambda_j$ measures the discriminative power of evidence $j$ to trait $\theta$; and $\varepsilon_{ij}$ is the Gaussian distributed error uncorrelated with $\theta$.

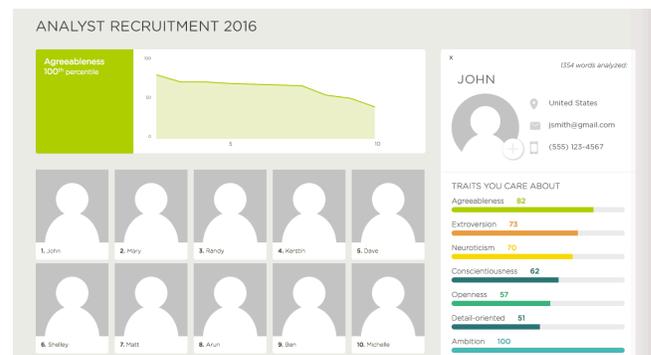

**Figure 3. A sample interview result showing candidates ranked by REP-inferred Big 5 trait "Agreeableness".**

To estimate the parameters in Formula 1, we use the conditional distribution of $X$ for trait $\theta$:

$$f(X_j|\theta_i) = \frac{1}{\sigma_j\sqrt{2\pi}} \frac{1}{X_j(1-X_j)} e^{-\frac{1}{2}\left[\frac{\ln\left(\frac{X_j}{1-X_j}\right)-\acute{\mu}}{\sigma_j}\right]^2} \quad (2)$$

Here $\acute{\mu}$ is the expectation of $\mu$ after a logit transformation, and $\sigma_j^2$ is the variance of residuals. We estimate the parameters using an EM algorithm over the training data [9].

*Step 3: Evidence-based Inference of Trait Score*
Given a user's text, this step analyzes the text and finds all matching linguistic evidence (e.g., matched words and phrases). The matching evidence is then used to compute a trait score by applying the trained model (Formula 1). Currently, we focus on inferring Big 5 personality traits and their 30 associated facets [10]. In our field deployment described below, only a user's chat text during the interview is used to infer his/her traits.

**Topic-based Conversation Engine**
To power virtual interviewers for practical, real world uses, our conversation engine must meet three design criteria. First, it must support diverse conversations and easy customizations of such conversations for a wide variety of interview situations, ranging from job interviews to patient inquiries to customer surveys. Second, the engine must be easily extensible to accommodate new capabilities, e.g., extending a virtual interviewer's ability to coach a user. Third, the engine must allow easy integration of third-party conversational technologies, e.g., Google NLP functions.

With these design criteria in mind and inspired by cognitive architectures [24], we have developed a novel, topic-based conversation engine to support mixed-initiative conversations. As described below, our approach models a conversation as a set of topics, which are flexibly threaded together to drive a conversation flow. As a result, conversations can be easily customized to accommodate various interview situations through topic specification. Moreover, conversational capabilities can be easily extended, incorporating even third-party functions through topics.

*Conversation Modeled as Topics*
We model a conversation as a set of *topics*, each of which defines a specific subject of the conversation. Semantically, a topic is similar to a discourse segment [13]. Figure 4 shows two example topics in a job interview conversation: one soliciting certain information from a candidate (Figure 4a) and the other answering a candidate's questions regarding the position (Figure 4b). Each topic further consists of one or more *semantic units*, each of which handles a particular aspect of the topic. For example, topic *ask-user-intro* has one unit, while topic *answer-job-inquiry* has two.

Each semantic unit includes a pair: a trigger and a response. A *trigger* encodes one or more conditions under which a semantic unit may be activated during a conversation, while a *response* defines one or more corresponding REP actions. In Figure 4a, the condition of asking a user for a self-introduction is at the beginning of an interview (*trigger* T1). If this condition is met, the REP asks the candidate to make a self-introduction (*response* R1). As shown in topic *answer-job-inquiry*, a trigger may be defined as a pattern that matches a user's input, such as "*when a decision will be made*". We describe below how a user's input is matched with a pattern defined in a trigger.

For extensibility, both trigger and response can contain patterns of expressions and functions. For example, in R1 the function (*ask-question*) generates a question that asks a user to make a self-introduction. An interview question, such as *user-intro-q*, may be defined first (Figure 4c) and then incorporated into a conversation topic, which affords proper expressions depending on the conversational context. For example, Kaya may phrase a question differently than Albert does, to reflect their respective personality.

Optionally, a semantic unit may be associated with a *sub-topic* that lets a REP have a deeper conversation with a user on the topic. In Figure 4a, (*drill-down*) is a sub-topic, which lets the REP probe further, e.g., asking the user to say more about herself. Sub-topics may be recursively defined to support virtually arbitrary levels of conversations.

*Conversation Flow Driven by Topic Activation*
Given a conversation that consists of a set of topics, our conversation engine decides when and which topics to activate, which in turn drives the conversation flow. To make these decisions, the conversation engine checks various properties of each topic. For this purpose, each topic is associated with a set of properties.

**Initiator**. First, there are three types of topics by their initiator. *Proactive* topics cover the subjects initiated by a REP in a conversation. In a virtual interview, proactive topics are often defined for a REP to pose interview questions and collect needed information from an interviewee. *Reactive* topics, on the other hand, cover the subjects initiated by a user. For example, topic *ask-user-intro* is a proactive topic,

```
(deftopic ask-user-intro                (deftopic answer-job-inquiry
T1:  [(chat-begin)]                     T2:  [when make decision]
R1:  [(ask-question user-intro-q)]      R2:  ["We will make a decision…"]
     (drill-down)
)                                       T3:  [how many apply]
                                        R3:  [(answer-num-candidates)]
                                        )
        (a)                                        (b)

(question [                             (config {
 user-intro-q {                          :agenda [ask-user-intro
  :type :open-ended                        answer-job-inquiry …]
  :heading "Could you introduce yourself?" :sidetalk
 } ] )                                   [:unordered movie sports]
                                         …} )
        (c)                                        (d)
```

**Figure 4. Topic representation with semantic units highlighted in grey. Labels T1-T3 and R1-R3 are not part of the definition and added for the purpose of illustration.**

while *answer-job-inquiry* is a reactive one. We also support *mixed-initiative* topics, which may be initiated by a REP or by a user. For example, a small talk topic may be initiated by a user ("*Shall we continue?*"), or by a REP ("*We still have time, let's chat about your favorite food*").

**Structure**. Sub-topics can only be activated if their associated parent topics are activated.

**Importance**. Just like in a human-human conversation, not all topics are equally important. Currently, we support three types of topics, ranked by importance. There are *agenda* topics that must be activated during a conversation at a certain point of time. In a virtual interview, agenda topics normally include all proactive topics that a REP wants to bring up and discuss with a user. There are *sidetalk* topics that may or may not be activated during a conversation. These topics are mostly used to help make conversational transitions or fill the gaps during a conversation. For example, assume that a REP has finished asking all the interview questions and there is still time left. One of the side topics may be activated to let the REP continue with a user. *Error-handling* topics are only activated if errors or unexpected situations occur during a conversation.

**Temporal Order**. Similar to a human-human conversation, certain topics (e.g., asking a user to introduce himself) are discussed before others (e.g., the user's opinion on mobile strategy). A partial temporal order may be specified among defined topics to control the topic order. Topics may not be ordered at all to support an arbitrary flow among them.

To drive a conversation flow, the conversation engine examines all topics and their associated properties, and then decides which topic to activate. For example, the configuration file in Figure 4(d) specifies that both topics *ask-user-intro* and *answer-job-inquiry* are agenda topics and one should be activated before the other. In addition, there are two side talk topics, *movie* and *sports*, which may be activated in any order after the two agenda topics are discussed.

A topic becomes an activation candidate if one or more of its semantic units have matched triggers. If there are multiple candidates, the engine ranks them by their structure, importance, and temporal order. Specifically, it ranks an associated subtopic the highest, then any agenda topics by their temporal order if any, followed by side talk and error-handling topics. This ranking allows a conversation to go deeper *and* guarantees all agenda topics be discussed. By default, a semantic unit (e.g., asking a user to introduce himself) can be activated only once. After it is activated, the conversation about this aspect is considered done. However, certain semantic units are *reusable*. For example, all error-handling units are reusable, since they can be activated throughout a conversation to perform error handling whenever errors occur.

*Pattern-based User Input Interpretation*
Accurately interpreting a user's input is one of the biggest challenges in building a conversational system. Although our REP infers a user's traits based on observed linguistic cues (Formula 1) without the need to understand every user input precisely, it still must understand a user's input adequately to drive the interview process. To support the real-world use of a REP, we adopt a pattern-based approach to user input interpretation, since it has shown its effectiveness in many conversational systems [45, 47]. Each pattern is defined by a composition of tokens, strings, or regular expressions, which are recognized as states in a finite-state machine (FSM). For generalization purposes, where a pattern can match with many input variants, our engine automatically lemmatizes tokens and inserts wildcards. Patterns are compiled into optimized FSMs by minimizing the number of states [17]. The FSM compiler itself is optimized by using several novel pattern rewriting techniques, such as recursively factoring out common prefixes and suffixes, retracting a large list of alternatives into a single special token, and eliminating duplicated patterns. As a result, hundreds of thousands of complex patterns can be compiled within seconds, and our FSM runtime can process more than 100 million user input tokens per second on a mid-tier Macbook Pro Laptop.

*Template-based System Response Generation*
Currently, we use a template-based approach to generate a REP's utterances during a conversation. To make a REP sound more natural, a response template may include multiple alternatives. For example, a response template that allows a REP to utter positive confirmations may include options, such as "*Great job*" and "*Well done*." At run time, the conversation engine may select an option randomly or by certain criteria, e.g., linguistic cues reflecting a particular personality trait (e.g., *cheerfulness*).

**DESIGN OF A VIRTUAL INTERVIEWER**
To test our hypotheses, we developed two virtual interviewers, each with a distinct personality that is associated with effective human interviewers [19, 50].

**Two Virtual Personas**
Existing studies suggest two types of effective human interviewers: one with a warm, cheerful personality, effective at building rapport with interviewees, and the other with a serious, assertive personality, effective at requesting cooperation from interviewees [19, 50]. We thus designed and built two virtual personas.

*Appearance*
Since people often associate one's appearance with a particular personality [22, 36, 48], we created two static profile images, each representing a distinct persona. In particular, Kaya takes the form of a cheerful female, while Albert is a serious-looking male (Table 1). Our gender assignment is based on psychology studies, which show women in general are warmer and more sensitive in their personality, while men are more reserved and dominant [12, 46].

*Conversational Behavior*
In addition to appearance, an agent's conversational behavior exhibits its personality [4]. Specifically, we unveil the personality of a virtual interviewer through four types of conversational behavior associated with an interview task.

**Effective Inquiring**. Clinical studies show that interviewers can use different techniques to effectively draw out truthful information from interviewees. For example, an *affective strategy* helps an interviewer establish rapport with an interviewee, which then allows the interviewee to open up and act authentically [15]. On the other hand, *negative politeness* may be used to preserve the emotional distance between an interviewer and an interviewee, minimizing imposition on the interviewee [5]. Since the use of particular strategies or techniques reveals an interviewer's personality naturally, we match our virtual personas with different inquiring strategies or techniques. Specifically, aligning with her warm and cheerful personality, Kaya uses an affective strategy and positive politeness in a conversation, while Albert uses a cognitive strategy and negative politeness to reveal his reserved and serious personality.

**Effective Influencing**. Besides inquiring about information, an effective interviewer may also want to influence interviewees (e.g., encouraging them to think outside the box) [33]. To achieve this goal, we introduce influence operators [4], which help our virtual interviewers effectively guide interviewees to perform their best. To naturally exhibit an interviewer's personality, we assign different influence operators: Kaya uses empathetic and cooperative language, while Albert's language is reassuring and forgiving.

**Small Talk**. While small talk helps make a conversation more engaging [3], it may also block honest interaction [16]. To match their personalities, Kaya uses small talk to build rapport and show her friendly side, while Albert rarely chitchats to reflect his serious persona. Small talk is often part of an interview question. For example, Kaya comments on herself "*Everyone has a role model including me :) Mine is Margaret Hamilton,*" before asking a user about his.

**Linguistic Style**. Since one's linguistic style unveils one's personality [35, 38], we incorporate matching linguistic cues (e.g., words, phases, and punctuations) into our virtual interviewers' utterances. These linguistic cues are selected based on previous studies [21, 23, 49] and our own findings that establish the relationships between various linguistic cues and personality traits (Formula 1). For example, Kaya frequently uses first-person, affective expressions ("*I love romantic movies as they make me cry.*"), emoticons, and exclamations. In contrast, Albert uses third-person declarative, projective statements (*"It makes sense."*) or abbreviated replies (*"Understood."*). Moreover, Kaya uses questions and suggestions to show her friendliness, while Albert uses assertions and demands to display his seriousness. Table 1 summarizes the two REPs and their matched behavior.

### Interview Questions

In our field deployment, we helped a firm screen their candidates for an open position. The firm first provided us with a set of interview questions to inquire about a candidate's interests and skills pertinent to the job. We then put in additional questions to: (1) make the interview flow more naturally, (2) measure candidates' willingness to confide in and listen to a virtual interviewer, and (3) solicit candidates' perception and feedback of the virtual interviewer. The interview questions were organized into five parts.

**I. Introductory Section**. The first section included five open-ended questions, helping a REP build rapport with a user. It started by asking the user to introduce herself, and then chatted with her to collect additional information, such as her hobbies, favorite movie, and role model.

**II. Impression Management Questionnaire**. The second section included a 20-item impression management (IM) scale [34] to measure one's willingness to confide in a REP.

**III. Opinion Questions**. This section included two open-ended questions that solicited a user's opinions. Since these questions were controversial in nature, these discussions were designed to measure a user's willingness to confide in and listen to a REP (see more below).

**IV. About You**. This section included ten questions, asking a user more information about herself, such as her job preferences, skills and interests, and strengths and weaknesses. The REP also shared and discussed its analysis of a user, such as her personality traits, top strength, and top weakness. The REP then suggested certain actions, which were meant to measure users' willingness to listen to the REP.

**V. Post Interview Survey**. This section collected basic demographics from users and their rating of the REP on several aspects, from personality to performance.

*Measuring Users' Willingness to Confide in a REP*
Drawing out truthful information from interviewees is paramount to any successful interview, since such information helps both interviewers and interviewees better assess real life situations (e.g., the fitness between a job and a candidate) [33]. However, due to social desirability biases—especially in risk-bearing situations like job interviews—interviewees may not be completely truthful [34, 51]. While studies show that users are more willing to confide in machines [27], we want to further understand how the personality of a REP affects one's willingness to confide in the REP, especially in high-stake situations.

|  | **Kaya** | **Albert** |
|---|---|---|
| **Static Profile** | 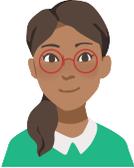 | 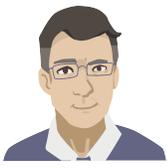 |
| **Personality** | Gregarious, Cheerful, Warm, Agreeable, Humorous; Like a friend | Reserved, Calm, Assertive, Rational, Careful; Like a counselor |
| **Effective Inquiring** | Affective strategy; Positive politeness | Cognitive strategy; Negative politeness |
| **Effective Influencing** | Empathy, Comfort, Frankness, Cooperation, Agreement | Reassurance, Commitment, Forgiveness |
| **Small Talk** | Personable | Minimal chitchatting |
| **Linguistic Style** | Questions, suggestions, affective expressions. | Assertions, projective statements, terse expressions. |

**Table 1. Two virtual personas and their behavior.**

Based on previous studies [2, 15, 27], we hypothesize that users are more willing to confide in a REP with a friendly, cheerful personality (Hypothesis I). To test our hypothesis, we assess one's willingness to confide in a REP by their willingness to share sensitive information with the hiring manager. In particular, we measure such willingness through two sets of interview questions: the Impression Management (IM) scale, and three sharing actions of revealing sensitive information about oneself or one's opinion on controversial topics with the recruiting firm.

The IM scale [34], also known as a "lie" or social desirability scale, is designed to detect how users consciously favor themselves to impress others. A user rates himself using a 7-point Likert scale on each of 20 items, such as "*I sometimes tell lies if I have to.*" The higher the score, the more inflation there is, and thus the less willing one is to confide in another with honest ratings.

In addition to the IM scale, we designed three user actions to measure one's willingness to confide in a REP. Each action is related to sharing with the hiring manager certain sensitive information or an opinion of a controversial subject. Table 2 summarizes these three actions. The first was to share a user's biggest weakness. The REP asked a user about her biggest weakness before showing its inferred top weakness for the user. The REP then asked the user to rate the inferred weakness by a *rating* in Table 2. It then asked the user to choose one of three actions (*sAction* in Table 2). The user's rating and choice of action were used to compute the user's willingness to confide in a REP (*A* in Table 2). By this measure, a user is most willing to confide in the REP if she chooses to share her biggest weakness suggested by the REP even if she *disagrees* with its analysis.

Two additional sharing actions were on controversial subjects (e.g., "*What important truth do very few people agree with you on?*"). For each subject, the REP first asked a user's opinion and then probed further by eliciting a rationale or introducing counter viewpoints. The REP asked the user how confident she was (*CF* in Table 2) before asking her whether she still wanted to share her opinion with the hiring manager (*sAction* in Table 2). Based on the user's reported confidence and choice of action, the willingness to confide in is then computed (Table 2). By our measure, a user is most willing to confide in a REP if she decides to share her opinions with the hiring manager despite having no confidence. Our rationale behind this measure is that users always want to show their best qualities in an interview.

| Share Weakness | $A = rating \times sAction$ <br> $rating = \begin{cases} 3, disagree \\ 2, not\ sure \\ 1, agree \end{cases}$ $sAction = \begin{cases} 0, don't\ share \\ 1, share\ mine \\ 2, share\ REP\ sugg. \end{cases}$ |
|---|---|
| Share Opinions ($j$=1, 2) | $B_j = CF_j \times sAction_j$ <br> $CF_j = \begin{cases} 1, High \\ 2, Med \\ 3, Low \end{cases}$ $sAction_j = \begin{cases} 0, don't\ share \\ 1, share \end{cases}$ |
| **Willingness to confide**: $WC = A + \sum_{j=1}^{2} B_j$ | |

Table 2. Measuring willingness to confide in a REP.

When they have an option not to share something they are uncertain of, they could hold back without risking to expose their potential weaknesses (unless they are willing to confide in a REP despite the risks).

*Measuring Users' Willingness to Listen to a REP*
In addition to eliciting authentic responses from interviewees, an interviewer should also effectively guide the interviewees and help them reveal their best qualities (e.g., demonstrating thoughtfulness). However, users may not always be willing to listen to machines [36, 37]. To test our hypothesis II, where users may be more willing to listen to a REP with a serious personality [19, 36], we designed a set of user actions to measure one's willingness to listen to a REP. Table 3 summarizes the actions and their measures.

In the first two actions, *willingness to listen* is measured by a user's click-through after a REP asked the user to read online articles related to the current interview topic by clicking on the articles' URLs.

Another five actions were related to persuading a user to share a piece of REP-suggested information. Similar to the classic Desert Survival Problem that is used to test the power of persuasion [37], all our five actions on testing one's willingness to listen had four steps. The REP first asked a user's input on a specific subject (e.g., her top strength) before offering a suggestion (e.g., the REP-inferred top strength), which was often different from the user's input. The REP then asked the user to rate its suggestion (*rating* in Table 3) before persuading her to share its suggestion, as opposed to the self-reported one, with the hiring manager. In these five actions, *willingness to listen* is measured by how strongly a user disagreed with a REP's suggestion and the user's final choice of action.

| Click-through ($j$=1, 2) | $C_j = \begin{cases} 0, URL\ not\ clicked \\ 1, URL\ clicked \end{cases}$ |
|---|---|
| Share Action ($j$=1, 5) | $S_j = rating \times act$ <br> $rating = \begin{cases} 3, disagree \\ 2, not\ sure \\ 1, agree \end{cases}$ $act = \begin{cases} 1, share\ REP's \\ 0, share\ mine \end{cases}$ |
| **Willingness to listen**: $WL = \sum_{j=1}^{2} C_j + \sum_{j=1}^{5} S_j$ | |

Table 3. Measuring willingness to listen to a REP.

*Measuring Users' Perceptions of a REP*

Based on previous studies on anthropomorphic interfaces [25, 36, 39, 40] and our hypotheses, we designed three sets of Likert-scale questions to measure users' perceptions of a REP. The first set asked users' perceived trustworthiness of a REP, since such perceptions may be directly related to their willingness to confide in and listen to the REP. To validate the effectiveness of our personality manipulation (i.e., whether users perceived a REP's given personality), the second set asked users' perceptions of a REP's personality from multiple aspects, such as perceived characteristics (e.g., how *cheerful* or *reserved* a REP is), role similarity (i.e., how a REP is similar to a friend or a counselor), and personality similarity (i.e., how similar a user's personality is to a REP's). The third set solicited user's feedback on a REP's usability (e.g., how enjoyable a REP is) [37].

## FIELD DEPLOYMENT

We deployed Kaya and Albert to aid a firm in hiring an associate. The firm notified all its 800 applicants to expect an email invitation of a virtual interview opportunity and encouraged them to use this opportunity to showcase their unique characteristics. Half of the candidates were invited to interview with Kaya, and the other half to interview with Albert. Each email invitation contained an individualized URL to start the interview at our website.

The process lasted about three weeks, during which 157 applicants completed their virtual interviews with Kaya, 159 completed theirs with Albert, and 59 people started but did not finish their interviews. Among the 157 users with Kaya, there were 130 (83%) male and 27 female (17%); among the 159 users with Albert, there were 125 (79%) male and 33 (21%) female. Users' ages ranged from 18 to over 50, and 90% of them were between 18 to 35 years old. On average, a user spent well over an hour with a REP and contributed about 1600 words.

## RESULTS

We first present an overview of the results and then our analysis around our three hypotheses.

### Overview of Results

After each interview, we asked users to rate their perception of the REP's personality traits (on a scale 1-5). The top three rated personality traits for Kaya were: *calm*=3.89, *cheerful*=3.875, and *warm*=3.82, and for Albert they were: *calm*=4.15, *rational*=3.96 and *assertive*=3.58. Except for the highly rated *calm* for Kaya, users' perceptions matched with the intended personality of Kaya and Albert (Table 1).

Table 4 summarizes various user ratings. An independent samples t-test showed no significant differences in most of user ratings except that Kaya was significantly more like a friend than Albert $t(307)=2.54$, $p<.01$. Each REP was also rated on three trust measures (integrity, competence, and benevolence) [31], which were combined additively to create an index measure called *Trust*. An independent samples t-test showed a strong trend that users trusted Albert more than Kaya, $t(314)=1.88$, $p<.06$.

| impression of REP | Kaya | Albert |
|---|---|---|
| Like a friend* | mean=2.74, sd=1.20 | mean=2.39, sd=1.21 |
| Like a counselor | mean=3.15, sd=1.24 | mean=3.20, sd=1.29 |
| Helpful | mean=3.63, sd=1.14 | mean=3.61, sd=1.01 |
| Likable | mean=3.54, sd=1.12 | mean=3.32, sd=1.03 |
| Insightful | mean=3.49, sd=1.16 | mean=3.58, sd=0.99 |
| Enjoyable | mean=3.56, sd=1.19 | mean=3.42, sd=1.15 |
| Interesting | mean=3.83, sd=1.24 | mean=3.91, sd=1.05 |
| Trust (p<0.06) | mean=3.65, std=0.99 | mean=3.84, sd=0.82 |

Table 4. Results of user ratings on REP (*p<.05).

### H1: User's Willingness to Confide in a REP

Our first hypothesis stated that users would be more willing to confide in Kaya as she should be better at building rapport. As mentioned earlier, we measured one's willingness to confide in a REP from two aspects: the IM Scale [34] and three actions measured by an index variable *Willingness to Confide* ( Table 2).

By the IM Scale, users were less likely to want to impress Albert (mean=12.68, sd=3.48) than Kaya (mean=12.93, sd=3.44). Using the IM scale as a dependent variable, we ran a General Linear Model (GLM) with three independent variables: *Trust* (user perceived trustworthiness of a REP), *Similar Personality* (how similar users felt the REP's personality was to their own), and *Agent* (distinguishing Kaya from Albert), controlling for user's *Gender*. The result showed *Trust* was the only significant factor: $F(1, 302)=5.65$, $p<.02$, *Coeff*=.59. This implies that the more users trust a REP, the more likely they want to consciously impress the REP. This finding seems consistent with the theory that people apply the same social rules to computers as they apply to humans [37]. Specifically, people want to impress someone whom they trust (e.g., parents).

With *Willingness to Confide* as a dependent variable, we ran a GLM with the same three independent variables: *Trust*, *Similar Personality*, and *Agent*, also controlling for *Gender*. It showed that *Agent* was significant: $F(1, 302)=5.38, p<.02, Coeff=.46$. This shows that users are more willing to confide in Albert *and* they are more willing to do so if they trust the REP more. *Gender* was not significant, indicating that the gender of the user did not influence their willingness to confide in a REP.

Our results above show that users are more honest (per the IM) and more willing to confide in Albert, *contrary* to our hypothesis that users would do so more with Kaya.

### H2: User's Willingness to Listen to a REP

Our second hypothesis assumed that users would be more willing to listen to Albert, who should be better at establishing authority. Using an additive index of users' actions, *Willingness to Listen* (Table 3) as the dependent variable, we ran a GLM over the same independent variables: *Trust*, *Similar Personality*, and *Agent*, controlling for *Gender*. It showed that *Agent* was significant: $F(1, 302)=4.15, p<.04,$

**Factor 1**: Emotionality, Anxiety, Vulnerability
**Factor 2**: Extraversion ← Cheerfulness
**Factor 3**: Neuroticism ← Depression
**Factor 4**: Self-Discipline → Conscientiousness ← Achievement-Striving,
**Factor 5**: Adventurousness → Openness
**Factor 6**: Intellect, Liberalism
**Factor 7**: Agreeableness ← Trust

Table 5. Personality traits loaded onto separate factors. The arrow indicates one trait is a facet of a Big 5 factor.

*Coeff=.39*. This result showed that users are more willing to listen to Albert, a REP with a reserved, assertive personality. However, since *Similar Personality* was not significant ($p<.09$), users may not necessarily be willing to listen to a REP who possesses a similar personality, in contrast to certain previous findings [37, 40].

### H3: Effects of Individual Differences on User Behavior

In our third hypothesis, we expected that users' individual differences, i.e., their traits inferred from their text during a virtual interview, would affect their perceptions of and behavior with a REP. A factor analysis was done first on the 35 inferred Big 5 personality traits [10]. We used a Varimax rotation with a Kaiser normalization. A scree plot[1] revealed that seven factors should be used, accounting for 50.5% of the variance (Table 5). To test our hypothesis, we used these factors as independent variables in regression models below.

*Personality Traits and User Perception of a REP*

We analyzed how the inferred users' personality traits affected their perception of a REP. We focused on one user perception variable that was more different for the two REPs (Table 4) and also most relevant to a virtual interview context: *Trust*. For each REP, we conducted a stepwise regression using *Trust* as the dependent variable and the seven factors of personality traits as independent variables. For Albert, *Conscientiousness* significantly impacted *Trust* whereas for Kaya, *Conscientiousness and Openness* significantly influenced *Trust*. This implies that conscientious users trust a REP in general, and with also open-mindedness, users will trust a cheerful REP like Kaya.

*Personality Traits and User Behavior with a REP*

We next examined how the users' personality traits influence their willingness to confide in and listen to a REP. Table 6 summarizes all the results.

For each REP, we conducted a stepwise regression with *Willingness to Confide* as the dependent variable and the seven factors as independent variables. For Kaya, *Emotionality* was significant, showing that more emotional, worried, and vulnerable users are *less* willing to confide in Kaya. No significant trait factors were found for Albert.

We then conducted a stepwise regression with the *IM* scale as the dependent variable and the seven factors in Table 5

---
[1] A scree plot is used to determine the number of factors to select based on a visual analysis of when the curve flattens.

---

as independent variables. *Extraversion and Neuroticism* were significant for Albert while no significant traits were found for Kaya. As a higher value of IM indicates a greater tendency to inflate one's own image, the results show an inverse relationship: extroverted and emotional users are less likely to inflate themselves to impress a serious REP.

Finally, we conducted a stepwise regression with *Willingness to Listen* as the dependent variable, and the seven factors as independent variables. *Agreeableness* was significant for Kaya implying that more agreeable and trusting users are more willing to listen to a warm and cheerful persona like Kaya. No trait factors were significant for Albert.

### Summary of Findings

We summarize three key findings and their implications on designing effective virtual interviewers. First, our field deployment demonstrated several practical values of a virtual interviewer. It makes a recruiting process more efficient, objective, and inclusive. Out of 800 applicants with half of them who simply ignored the virtual interview invite and 59 who did not finish their virtual interview, the hiring manager was able to quickly identify 12 quality candidates out of the 316 who completed their virtual interview. He also commented that a few of 12 would have gone unnoticed by judging their resumes alone. Moreover, users appear to act authentically around a virtual interviewer to reveal their true character. For example, one told Albert that his question was dumb, while another told Kaya to "drink warm water" after she said "I just had a brain freeze" when recovering from an error. Users also "*felt more casual speaking with Kaya than an actual person, almost like writing in a journal*", while others found the REP's analysis very insightful, e.g., "*It was not something I had thought about before and appreciate the information*".

Second, creating a virtual interviewer with Albert's personality and its matching behavior (Table 1) will make users willing to confide in and listen to the interviewer.

Third, it is desirable to customize a virtual interviewer's behavior based on a user's inferred traits to make an interview more valuable. For example, our results show that achievement-striving users tend to trust a REP more and such trust also makes them eager to impress the REP. In such a case, a REP can remind such users to be honest, be-cause they can achieve more with their true selves.

| Variable | REP | Sign Factor | Coeff | F (df) | p |
|---|---|---|---|---|---|
| Perceived Trust | Albert | Conscientiousness | .17 | F(1, 153)=6.83 | .01 |
| | Kaya | Conscientiousness Openness | .15 .25 | F(1, 154)=7.58 | .001 |
| Willingness to Confide | Albert | n.s. | | | |
| | Kaya | Emotionality | -.26 | F(1, 155)=4.19 | .04 |
| IM scale | Albert | Extraversion Neuroticism | -.59 -.63 | F(2, 152)=4.87 | .009 |
| | Kaya | n.s. | | | |
| Willingness to Listen | Albert | n.s. | | | |
| | Kaya | Agreeableness | .28 | F(1, 155)=4.97 | .03 |

Table 6. Personality factors predicting user behavior.

## DISCUSSION
Our current work has several limitations that we discuss below, along with ongoing research, including design implications on building hyper-personalized intelligent agents based on automatically inferred user traits.

### Experimental Constraints
While we have tested our hypotheses in real-world job interviews, this context posed a couple of limitations on the understanding of our results. First, our users were real job seekers who had much at stake; thus the risk-bearing situation itself might have influenced their interview behavior. It is unknown whether users would behave similarly in a less risky context, such as routine patient interviews or customer surveys. Second, while previous work shows that gender impacts user interaction with agents [39], we did not find gender be a factor. As we are deploying REPs to support diverse interview situations, it would be interesting to see whether our findings would hold across interview contexts.

### Customizing an Interview
Currently, each interview question is represented by a conversation topic, and follow-up questions are scripted as sub-topics. While it is straightforward to customize an interview by adding or removing interview questions as topics, it is impractical for a business user, such as a hiring manager, to do so manually. The task becomes even harder if an interview question is an open-ended question, which would then require a business user to consider possible user responses and how to handle them as sub-topics. Ideally, a business user needs to enter just her interview questions, which are then automatically translated into conversation topics. Toward this goal, we are examining patterns in interview questions before auto-translating them into topics.

### Inferring User Traits Beyond Text
Currently, a REP infers a user's traits from the user's text given during the interview. However, we observed interesting user behavior beyond what their text captures during the whole process. For example, certain users asked for help (via email) when encountering technical difficulties (e.g., using an incompatible web browser) while others simply gave up (e.g., some abandoned partially completed interviews). Moreover, during an interview, a user may respond to a question promptly but hesitate to reply to another question. In addition to a user's text, all such behavior could have been leveraged to build a better trait inference model. The challenge is how to capture such behavior, especially outside of a virtual interview, and how to relate user behavior with specific traits.

### Validating Inferred Traits
Since a unique capability of our REP is to assess a user's traits automatically, we have evaluated our trait inference engine in several ways. First, we used our model to analyze over 15 million Twitter and Facebook users. We randomly selected 100,000 of them to evaluate the reliability of our model by computing the Cronbach's $\alpha$, a metric estimating the reliability of a psychometric test [7]. Our results showed that our model requires about 1000 words to achieve a good reliability ($\alpha >= 0.8$) for many traits (e.g., 19 of 35 Big 5 personality traits). Second, our current field deployment demonstrated the effectiveness of our model, since the hiring manager was able to quickly find matching candidates based on the ranking of their traits. However, completely validating such a model is non-trivial [14] and requires further work. One direction we are pursuing is to examine how well the inferred traits predict users' real-world behavior, such as job performance.

### Design Implications Beyond Virtual Interviewers
Our findings indicate that inferred user traits significantly influence users' behavior with a REP. This implies that a REP can dynamically adapt its behavior to a user based on the user's traits inferred during a conversation. For example, when an *impulsive* user is unwilling to listen to a REP's guidance during a coaching session, the REP would try to calm him down first before continuing. On the other hand, a REP would encourage a *humble* person to open up and show her best talents. While much research recognizes the importance of incorporating user traits, such as personality, into an intelligent system [18, 42], few practical systems have been developed to obtain and use such traits in real time. Thus, automatically inferring user traits during a human-computer conversation enables the development of a new generation of intelligent agents, who can truly understand their users on the fly and adapt to them in real time.

## CONCLUSIONS
We have presented an intelligent virtual interviewer, called a REP, who interviews a user through a text-based conversation and automatically infers the user's traits, such as personality and strengths. In this paper, we show how two REPs with distinctly different personalities influence users' behavior—their willingness to confide in and listen to a REP during an interview. Our results from a field deployment of the two REPs with 316 completed interviews revealed several new findings. First, a virtual interviewer can make a recruiting process more efficient, objective, and inclusive. Second, users act authentically around a virtual interviewer to reveal their true character. In particular, users are more willing to confide in and listen to a serious, assertive REP. Third, users' personality traits influence their perception of and behavior with a REP. It thus is highly valuable to create hyper-personalized REPs based on users' traits that are automatically inferred during a conversation.

## ACKNOWLEDGEMENTS
This work is funded in part by the Air Force Office of Scientific Research under FA9550-15-C-0032. The authors would also like to thank Frances Thai and Connie Truong for their help in developing the dashboard.